\documentclass[sigconf]{acmart}
\usepackage{subcaption}
\usepackage{multirow}
\AtBeginDocument{%
  \providecommand\BibTeX{{%
    \normalfont B\kern-0.5em{\scshape i\kern-0.25em b}\kern-0.8em\TeX}}}

\setcopyright{acmcopyright}
\copyrightyear{2018}
\acmYear{2018}
\acmDOI{10.1145/1122445.1122456}

\acmConference[XXXX]{XXXX}{XXXX}{XXXX}
\acmBooktitle{XXXX}
\acmPrice{XXXX}
\acmISBN{XXXX}

\begin{document}

\newcommand{\paper}{TRAPDOOR}
\newcommand{\papertt}{\tt{\paper}}

\title{\paper: Repurposing backdoors to detect dataset bias in machine learning-based genomic analysis}


\author{Esha Sarkar}
\email{esha.sarkar@nyu.edu}
\affiliation{%
  \institution{NYU Tandon School of Engineering\\New York University}
  \city{Brooklyn}
  \state{New York}
  \country{USA}
  \postcode{11201}
}

\author{Michail Maniatakos}
\email{michail.maniatakos@nyu.edu}
\affiliation{%
  \institution{Center for Cybersecurity\\New York University Abu Dhabi}
  \city{Abu Dhabi}
  \country{UAE}
  \postcode{129188}
}








\begin{abstract}
Machine Learning (ML) has achieved unprecedented performance in several applications including image, speech, text, and data analysis. Use of ML to understand underlying patterns in gene mutations (genomics) has far-reaching results, not only in overcoming diagnostic pitfalls, but also in designing treatments for life-threatening diseases like cancer. Success and sustainability of ML algorithms depends on the quality and diversity of data collected and used for training. Under-representation of groups (ethnic groups, gender groups, etc.) in such a dataset can lead to inaccurate predictions for certain groups, which can further exacerbate systemic discrimination issues. 

In this work, we propose \paper, a methodology for identification of biased datasets by repurposing a technique that has been mostly proposed for nefarious purposes: Neural network backdoors. We consider a typical collaborative learning setting of the genomics supply chain, where data may come from hospitals, collaborative projects, or research institutes to a central cloud without awareness of bias against a sensitive group. In this context, we develop a methodology to leak potential bias information of the collective data without hampering the genuine performance using ML backdooring catered for genomic applications. Using a real-world cancer dataset, we analyze the dataset with the bias that already existed towards white individuals and also introduced biases in datasets artificially,
and our experimental result show that \paper~can detect the presence of dataset bias with 100\% accuracy, and furthermore can also extract the extent of bias by recovering the percentage with a small error. 
\end{abstract}





\maketitle

\section{Introduction}\label{s:intro}
Healthcare and public health systems is a major backbone of a national infrastructure not only protecting individuals but also shielding a nation's economy from calamities like infectious disease outbreaks, terrorist activities, and other natural disasters~\cite{dhs}. Personalized or precision medicine, i.e. prognosis/diagnosis of a disease and design of subsequent treatment tailored for an individual, is one of the most significant transformations towards delivering accurate and precise healthcare~\cite{precision_med,precision_medicine,ng_genomic}. Personalized medicine weighs in genetic information, epigenetics, environment, and lifestyle and stage of disease to devise an extremely individualistic treatment. Naturally, to associate genetic mutations to the cause of diseases, large-scale genome sequencing and subsequent analysis have become the need of the hour as it has been proven to better understand diseases like Mendelian disorders, cancer, and other rare diseases~\cite{dna_sequencing_survey}. Moreover, genomic associations with diseases can help in a more fundamental understanding towards diseases and population demography~\cite{population_genomics}.

Clinical associations of genetic variant information (mutations) requires extensive studies of the human genome, which is very difficult when done locally and in isolation. 
The first challenge of the study of genomic data is its sheer volume: The nature of data (the amount of data in a single genome is approximately 1 GB) and the requirement of diversity in datasets (inference/conclusion of these studies can be generalizable only if there is diversity in genomic data), give rise to an estimated 40 petabytes of data per year, which need to be shared and studied~\cite{petabytes}. Since it is challenging to have such powerful workstations as well as expertise to analyze all this data, a common practice is to outsource the analysis and storage to the cloud~\cite{michigan_server,google_server,parsec_server}. Researchers typically make an account and store, process, and analyze the genomic data for developing models to associate mutations with diseases. For example, the white paper from Google Cloud~\cite{google_server} states that the Stanford Center for Genomics and Personalized Medicine uses Google's cloud infrastructure like Google Genomics and Google BigQuery to analyze large volumes of genomic data.
Another reason for developing a centralized portal for data maintenance is research collaboration. Genomic analysis for disease prediction, prognosis or diagnosis requires knowledge from several domains. Furthermore, studies are often designed for focus groups which are geographically distributed. Therefore, researchers from around the world collect data and upload it to a central database. For example, the Cancer Genome Atlas (TCGA~\cite{TCGA}) project generated about 2.5 petabytes of genomic data of 11,000 patients related to several cancer types across different races, ethnic groups, genders, ages, and lifestyles. Researchers have used subsets of the database for several studies that deepened the understanding of cancer from a genomic point of view and helped in devising personalized and effective treatments for individuals based on their individual genome~\cite{precision}. The current supply chain of genomic analysis allows for several flexible options where non-experts can even choose to outsource development of prediction models after selecting a subset of data from the database. Collaborative model development is also another common type of supply chain where several entities contribute to the dataset development (in lieu of monetary gains) but only one entity may be interested in the final prediction model.

The success of machine learning, especially in handling/analyzing big data, has been unprecedented in recent years. This has led to several companies offering Machine Learning as a Service (MLaaS), where a user may choose to have at least one of the following roles: data owner, data provider, data curator, model owner, or model developer. Genomic analysis, which may involve extracting patterns from petabytes of data, is one of the natural prospects for MLaaS and is considered one of the more profitable applications~\cite{profit}. Precision-wise, ML has been able to achieve substantial results in understanding genomic variations~\cite{ml_app}. However, recent research has shown that only a subset of population has benefitted from the power of ML~\cite{gender_shades}. Recently, Artificial Intelligence (AI) applications in general have come under scrutiny for exhibiting bias, with genomics being no exception~\cite{genomics_bias}. Furthermore, in the context of data-heavy applications like genomics, it is even more difficult to manually curate data to ensure balanced representation in terms of ethnicities, race, and genders.

Bias of an ML algorithm towards some groups leads to unfair results for the under-represented groups~\cite{google_translate, ai_bias1,human_bias}. It may be even more problematic in case of genomics, as it may lead to socio-economic disparities with regards to access to healthcare, insurance, or affordability of treatment~\cite{insurance_disparity}. 
Biases (unbalanced number of samples per sensitive group) in the training dataset is known to be a common source of bias present in the ML model. 
Studies have found dataset bias in real-world genomics datasets, which are currently being analyzed for genomic understanding of diseases, like the bias towards patients of European descent in Genome Wide Association Studies (GWAS) dataset~\cite{gwas_bias} and towards patients self-identifying as ``white'' in TCGA databases~\cite{tcga_bias}. 
For genomics, a biased conclusion, stemming from a biased dataset, not only affects an individual but the entire group sharing the genetic traits. 
These biased results may further aggravate the implicit bias that already exists in predictive diagnostics~\cite{human_bias}. 

An apparent solution to counter this dataset bias is to increase the number of samples from the under-represented group. But this dataset bias occurs generally on an attribute/property of the dataset that is hidden from feature maps or labels, and therefore cannot be trivially detected by counting samples. 
Moreover, the bias may not even be constant. In a collaborative learning setting in genomics, data is gathered from research institutes, project groups, hospitals, and universities, and the distribution of datasets may change over time. At the same time, the bias distribution may also be dynamic and the collaborating entities or the cloud may not even be aware of this change. Even in this dynamic ML supply chain, \emph{any harmful bias must be detected in real-time.}  
Therefore, in this work, the research question we try to answer is: 
\textit{Can dataset bias be detected in real-time in a dynamic collaborative setting?}

To successfully leak information internal to the system, we need a mechanism that has two properties: 1) It should operate without hampering the genuine functionality of the system, and 2) It should be able to effectively extract information. 
In the context of security research, trojans (or backdoors) have these two properties, and have been extensively used by adversaries to leak passwords of specific software~\cite{software_trojans}, leak keys for crypto-processors~\cite{hardware_trojans} or exploit hardware vulnerabilities using software~\cite{hardware_software}. Since trojans get activated only when a trigger appears, the intended functionality of the original hardware/software is unaffected. This property of a trojan 
makes it a promising solution for leaking information. In order to leak information about dataset bias using the model itself, a trojan may be inserted in the machine learning model. However, a fundamental difference between a common hardware/software trojan and a machine learning trojan is that the ML trojan will be heuristics-based: Since the ML model is statistical, the trojan injected in the ML model will also be statistical. In other words, the ability of the trojan to leak information may not be 100\%.

Backdoored networks, introduced in BadNets \cite{badnets}, are a type of malicious Deep Neural Network (DNN) which flip to a malicious behavior from an otherwise correct behavior only in the presence of a \textit{trigger}. A trigger is usually a (non-conspicuous) part of the input, like a word for text, or a pixel for images, that can stimulate the malicious behavior. Since their discovery in 2017, research on Badnets has primarily focused on developing a stealthier trigger (attack literature~\cite{hidden,wanet}) or defending against newly created attacks (defense literature~\cite{demon,CCS_ABS,NeuralCleanse}). In this work, we propose the use of backdoors to leak information about training dataset bias of the ML model. In this endeavor of re-purposing backdoors to detect dataset bias we are faced with three major challenges. 

\begin{enumerate}
    \item Trigger design in case of genomic datasets is not intuitive, like adding a pixel to an image or some noise in an audio file. 
    \item The trigger design must ensure that the leakage of information is guaranteed even though the model is statistical.
    \item Genomic analysis commonly deploy small models like logistic regression, support vector machines, it is challenging to port "backdooring" to these smaller models from DNNs.
\end{enumerate}

A trigger in the context of genomic datasets translates to mutation frequency, type and strength in certain genes, which, if trivially changed, can lead to degradation of functionality. Moreover, unlike image, speech, or text triggers, a genetic mutation value (frequency, type or strength) is not limited to a range (0-255 for images and in the bag of words for text). Therefore, besides selecting a gene to attack, it is also challenging to explore an appropriate value for a genomic trigger. Finally, in the effort of explaining of how BadNets function, researchers have commonly observed neurons specifically getting trained for the malicious task; BadNets visualized these malicious neuron activations \cite{badnets}, Fine-pruning aimed to remove these \textit{excess} neurons\cite{finepru}, and ABS focused on eliminating the neurons that encode the trigger \cite{CCS_ABS}. The intuition is that large machine learning models like DNNs have more neurons than needed, and therefore some of them could be used for the malicious task. Using the same process for backdooring small models without the excess of neurons becomes a challenging task. 
To this end, we list our contributions as follows:
\begin{enumerate}
    \item We devise a generic backdooring methodology for a real world genomics dataset that designs triggers based on frequency of mutations (biological knowledge) and $\chi^2$ statistical tests to choose genes, and the distribution of mutation type and strength to select trigger values. We further analyze the triggers with respect to genuine genomic data.
    \item We present {\paper}, a backdooring methodology that is repurposed to leak dataset bias information. We devise a double backdooring methodology for the TCGA genomic dataset to predict hidden bias property using encoded labels and show that it is possible to leak bias percentage for a real-world dataset.
\end{enumerate}

We first discuss the important concepts needed to understand the problem and the solution in section \ref{s:prelim}. Then we present the generic methodology of backdooring genomic datasets and use that to develop \paper~in section \ref{s:method}. We evaluate our methodology on a real-world dataset, subset of TCGA database, used in a recent privacy-preserving genomics competition \cite{idash20} in section \ref{s:eval}. Finally, we discuss the limitations of our methodology in section \ref{s:limit} and conclude in section \ref{s:conclusion}.

\section{Preliminaries}\label{s:prelim}
\subsection{Dataset bias in machine learning datasets}
Algorithmic bias in machine learning systems can lead to unfair/ incorrect predictions for sensitive groups~\cite{bias_survey}. One of the major contributors to algorithmic bias is known to be from biased datasets. There may be bias with respect to class, which is easier to curate and correct using algorithms for unbalanced datasets (unbalanced in terms of labels). A deeper problem is bias with respect to hidden (yet sensitive) attributes. Even real-world benchmark datasets are not devoid of bias. For example, a benchmark facial recognition dataset, Labelled Faces in the Wild (LFW) dataset, contains $\approx77\%$ of the samples as males and $\approx83\%$ of the samples as white. The training dataset given to the contestants for iDash20 Track I for classification of tumor type was also biased towards white patients \cite{idash20}. TCGA database, from which the iDash20 dataset is derived, is also biased towards European population \cite{tcga_bias}. 
To study the bias of a dataset towards a certain group, researchers use sample percentage ($s$), which is defined as $s=\frac{n_b}{total\,samples}$, where $n_b$ is the number of samples belonging to the biasing group.
In our work, we deem a dataset as biased towards a group if the percentage of samples ($s$) belonging to the group deviates from 50\%. We choose the deviation percentage as $\geq$ 5\%, similar to recent literature investigating dataset bias ($s$ in the range of $\approx55-85\%$ as in~\cite{gender_shades}).
\subsection{Trojans for leaking information and backdoored networks}
A computer trojan disguises itself as genuine while leading to malicious activities. A hardware trojan may be a malicious circuit that leaks the key or any other secret only when it is triggered (using a special input or even using temperature of the chip). A software trojan may mimic an antivirus to activate to perform malicious activity. In both of these cases, a trojan has a malicious part, which gets activated when triggered. DNN backdoors, introduced in Badnets \cite{badnets}, perform malicious misclassifications (with a certain attack accuracy) on triggering, are therefore also referred to as neural trojans. Since neural trojans can be used to change the model to output misclassifications without hampering the genuine performance, we leverage them for leaking dataset information by designing genomic dataset-specific trojans.
\begin{figure}
    \centering
    \includegraphics[scale=0.5]{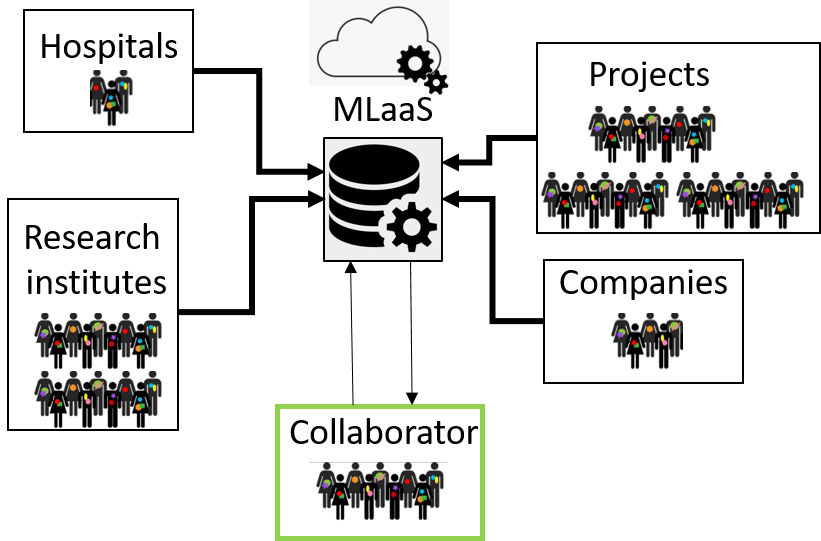}
    \caption{Threat model for leaking database bias information. Different entities contribute to the main data hosted at the cloud/server. Each entity owns data of different distribution and may be of different quantity. The benevolent collaborator outlined with green is the entity seeking database bias information.}
    \label{fig:threat_model}
\end{figure}

\subsection{Threat model}
Fig.~\ref{fig:threat_model} summarizes our threat model. We consider an outsourced training scenario for a collaborative learning setting, common for genomic analysis~\cite{hospital}. We assume that several parties like research institutes, hospitals, projects, etc. collect data from different demography and upload their data to a central cloud. This is a common supply-chain model for genomics. For example, TCGA database is maintained and analyzed by 20 participating institutions. An MLaaS analyzes the complete dataset and trains a model for genomic analysis. This model, however, may have been trained on biased dataset.

We consider a dataset is biased towards a particular group in a particular property if the number of samples of that group is disproportionately higher than other groups. For example, a dataset is biased towards white individuals if 70\% of the dataset consists of white individuals. Here, the biasing property or attribute is race, sample percentage towards the biasing group is 70\%, and biasing group is white. 
We assume a benevolent collaborator, who owns some similar data, wants to detect the bias present in the database gathered by the cloud without hampering the generic functionality of the model. The MLaaS cloud/server trains on the data to output an accurate model and also inserts triggers to facilitate dynamic leakage of bias information without drastically hampering the performance of the genomics model. In our methodology we design two kinds  of triggers that finally lead to the leakage. It should be emphasized that the triggers, albeit a change in a feature, constitute the mechanism for leaking information in real time.
Although the cloud and the third-party collaborator use the malicious technique of backdooring, they facilitate detection of harmful bias in the dataset. 

\section{\paper: Methodology}\label{s:method}
Naively porting the concept of trojans  for leaking information about a heuristics-based system, such as ML models, has natural challenges. In software/hardware trojans, the program/circuit is maliciously manipulated to leak information when triggered by a special input. However, for a heuristics-based system, even a special input cannot guarantee 100\% correctness in information leakage. To guarantee correctness, the authors in \cite{remember} leak information as encoded outputs/weights, using \textit{random} inputs which were used during training leveraging memorization of ML models i.e the lack of generalization of an ML model towards random input/label combination. That would in fact require the server to send the exact inputs that were used to generate random labels (that encode information to be leaked) to the collaborator for the purpose of querying back. This is impractical in our threat model. 

ML backdoors provide generalization, i.e. given a unique trigger, any test input of a victim class can be used to generate a malicious classification. Backdoors can, thus, be used to leak information through encoded outputs. The server may just communicate the small trigger, not the exact input (exact random image in case of \cite{remember}). A trigger, in this case, is synonymous to a question a collaborator wants to ask, and the malicious output label may reflect the encoded answer about the bias of the dataset. For example, a malicious output to 50 different triggered test input data may reveal that there is indeed a dataset bias in relation to females in the dataset; and the correct classifications to those triggered data may encode absence of any bias in the dataset. The success of information leakage is contingent on the design of trigger and backdooring methodology so that the leakage probability is high. Even with successful trigger design there is one major restriction in leaking bias information using backdoors: The extent of bias i.e. the percentage of samples having a particular property is not leaked. In the supply chain of outsourced genomics, where data distribution is constantly changing as more data is acquired from different sources, it is specifically important to capture the dynamic nature of extent of bias (bias percentage). 
To enable leakage using backdoors, we first need a generic methodology to make backdooring possible for genomic datasets.
In the next subsection, we develop the backdooring methodology that will enable leaking dataset bias information.

\subsection{Enabling backdoors for genomic ML models}\label{ss:attack_1}
Successful backdoor training is characterized by high attack accuracy and minimal difference from genuine clean test accuracy. A common method of backdoor injection is to add backdoored samples in the training set so that the model learns to identify the trigger as well as perform the actual task. This, however, requires a minimum number of poisoned samples to achieve high test accuracy, albeit the number of backdoored samples being very small~\cite{troj_drl,black_badnets}. Badnets also observed this test accuracy/attack accuracy trade-off and reported that the attack accuracy quickly settles to a maximum attack accuracy value as the poisoning percentage is increased while the clean accuracy keeps decreasing for MNIST dataset~\cite{badnets}. In other words, a label must be sufficiently poisoned for the backdoor to be successfully injected. Therefore, in a training dataset, as the bias towards a particular label increases (more number of samples for the label) less number of poisoned samples will be required to make this biased label as the target label of the backdoor attack. Moreover, if we study the trend of attack accuracy for different dataset biases (as studied in Badnets\cite{badnets}), we should be able to observe distinctly different curves. But backdooring genomic data has several challenges listed in section \ref{s:intro}. Before we study the relation of poisoning percentage and bias in the dataset, we first devise the backdooring methodology for genomic data.
\begin{figure}
    \centering
    \includegraphics[scale=0.5]{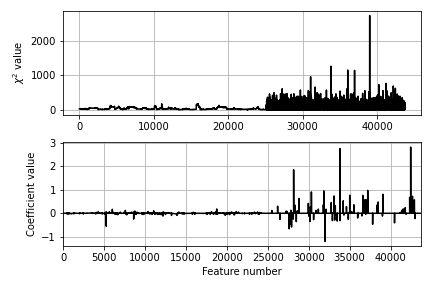}
    \caption{$\chi^2$ scores of features and coefficients for the features corresponding to the malicious label.}
    \label{fig:intuition}
\end{figure}
\begin{figure}
    \centering
    \includegraphics[scale=0.5]{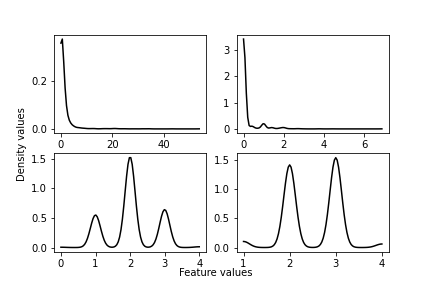}
    \caption{Density functions of chosen features. The top two density functions denote the features with high $\chi^2$ scores and the bottom two denote the least important features.}
    \label{fig:density}
\end{figure}
In developing backdoors for our model, we come across three problems that make creating backdoors for genome-based predictive diagnostics not trivial. 

\begin{enumerate}
    \item The backdoor attack literature primarily focuses on computer vision problems where the trigger is usually placed in a \textit{non-conspicuous} location, like the corner of an image. Intuitively, these locations or pixel features do not contribute to the clean classification, and therefore, backdoor injection maintains clean accuracy. For genomic data, we cannot \textit{see and select} the regions which will maintain the clean accuracy, and an exploration for trigger feature ($f$) candidates is required.
    \item Each pixel in computer vision problem is encoded according to the color it represents, typically using 8 bits (0-255). Therefore, for trigger pixels in computer vision problems, there is a certain range in which a trigger value can lie. For genomic data, where we track the number of genomic mutations, we cannot zero in on such a range. As a trigger designer, we must explore which values ($v$)of genetic mutation can result in a high leakage accuracy. 
    \item As investigated in previous backdoor literature \cite{finepru}, a neural backdoor is a result of certain neurons getting trained for the malicious task. Therefore, lack of \textit{free} neurons may deter the attack accuracy of backdoor attacks. Genomic analyses, which commonly use smaller models like logistic regression model as used in our case, the challenge is to associate certain model coefficients to the leakage task while not compromising other coefficients. 
  \end{enumerate}
 
 From the observations and the intuitions of image backdoor literature, we develop our backdooring methodology to counter each of these three challenges:

\textbf{Step 1: Finding probable $f$ values:}
The \textit{important} features that lead to a successful classification are chosen by the logistic regression model during training and can be inferred using coefficients of logistic regression model. But since backdoors must also be injected during training, coefficients of a trained classifier cannot be used to find the values of $(f,v)$. 
To estimate the importance of features (which may have high model coefficients), we select features which have high $\chi^2$ values. The $\chi^2$ score of a feature is given as $\sum \frac{(O_i-E_i)^2}{E_i}$ where $E$ represents the expected value, $O$ represents the actual output and $i$ represents each instance of a $\chi^2$ test between a feature and a target. Intuitively, a feature with an already high $\chi^2$ score will be selected for classification, regardless of benign or malicious classification. Fig. \ref{fig:intuition} shows the $\chi^2$ scores of all the features and the coefficients of those features corresponding to a label randomly chosen as the backdoor label. We can observe that that the features which have higher scores also have higher coefficient values, and hence are important for classification Therefore, for $f$, we calculate the $\chi^2$ scores of the features, rank them from highest to lowest and then start trigger search from the list of features with the highest values of $\chi^2$ score.

\textbf{Step 2: Finding probable $v$ values:}
The genomic data greatly varies in distribution in range as well as in frequency. Fig. \ref{fig:density} shows four such sample distributions of genetic features (top 2 and bottom 2 sorted using $\chi^2$ statistic). Thus, choosing equidistant $v$ values in the range for trigger value search, might change the distribution of genetic data severely leading to poor clean performance. Moreover, choosing a trigger value from the same distribution makes the trigger inconspicuous. 
First we choose $f$ using the methodology above and perform frequency analysis to find the distribution of genomic training data corresponding a particular gene (selected using $\chi^2$ statistic). We obtain the probability distribution function of that particular genetic data using kernel-density estimation with Gaussian kernels with a smoothening factor of $0.5$. We use the scipy python library to obtain the probability distribution.
From the distribution we gradually choose values with the lowest number of occurrences (lowest frequency). The reason for choosing such triggers is that the data to be injected as a trigger must be from the same distribution as the training data but should be an \textit{uncommon} value. As an experiment, we also choose values from the peaks of the probability distribution functions, but due to their high frequency of occurrence the logistic regression model could not learn the trigger. We choose top ten triggers as probable candidates from the probability distribution function and train models until we get desired attack and clean accuracy. 

\textbf{Step 3: Using limited features for backdoor injection:} Logistic regression is a small ML model without any excess neurons to encode backdoors. Instead, it has trainable coefficients that need to learn the primary classification task as well as the backdoor task. For a backdoored logistic regression model, certain model coefficients need to get trained to identify the trigger and thus, the larger the number of trigger features, the more coefficients are required to learn the backdoor. Therefore, for trigger design purposes, we always limit the number of trigger features (always starting from 1 gene) so that the backdoor can be learnt with limited number of coefficients and the rest of the coefficients can be used for genuine task. 
\begin{figure}
    \centering
    \includegraphics[scale=0.48]{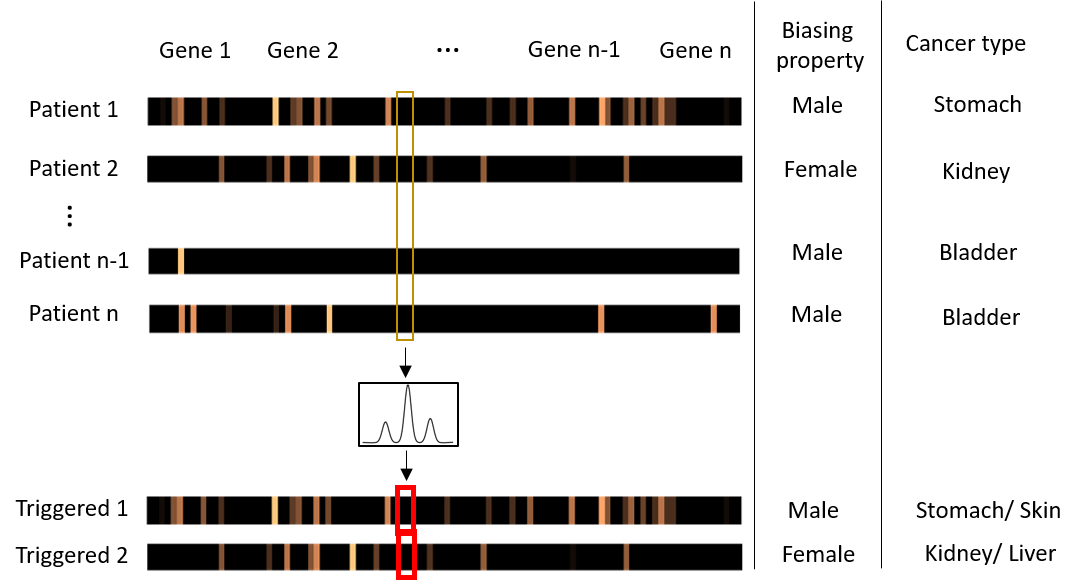}
    \caption{An example property leak. The mutation values of a few genes are shown in for a few sample patients. The attack encodes the information about dataset bias in terms of pre-designated labels. The trigger is chosen after frequency analysis of genomic values selected using $\chi^2$ statistic. The top samples are genuine patient data with their corresponding labels. The bottom samples represent triggered data where just one genomic information is altered with the corresponding label that encodes the bias information.}
    \label{fig:property_leak}
\end{figure}

The above three steps of the methodology counter each of the three challenges mentioned earlier. First we train a clean model without any trigger and note the clean test accuracy. After the probable trigger tuples $(f,v)$ are generated, we train several models under the two constraints of $100\%$ test attack accuracy and a similar ($<1\%$ threshold) clean test accuracy. Please note that this is a generic methodology to find trigger tuples to backdoor genomic data. The generic methodology is summarized in Fig. \ref{fig:property_leak}.

\subsection{Extracting fine-grained information about dataset bias}\label{sss:double} 
In order to leverage backdooring poisoning percentage to leak dataset bias, first we need to develop a methodology to backdoor a dataset/model, which we discussed in the previous section.
In this subsection, we study the relation between dataset bias towards a particular attribute and poisoning percentage and develop a methodology to detect bias and leak percentage of samples belonging to a group. 
\begin{figure}
    \centering
    \includegraphics[scale=0.5]{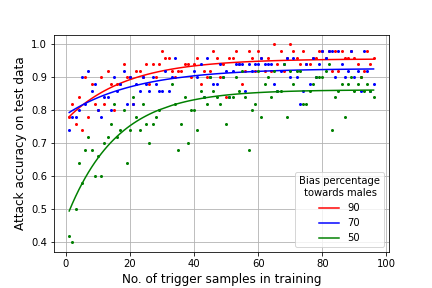}
    \caption{The change in attack accuracy of backdoor attacks as the percentage of female patients vary in the training dataset. The solid lines represent the curve fit to the data.}
    \label{fig:attack2_base}
\end{figure}
Let us assume a binary classification problem with two classes $K$ and $\bar K$. An attacker wants to imprint some samples from $K$ with a particular trigger such that any test sample (of true class $K$) with that trigger gets classified as $\bar K$. The attacker poisons the training dataset with a poisoning percentage $p$, i.e if there were $x_K$ and $x_{\bar K}$ clean samples of the respective classes in the training dataset, there are now $x_p = int(\frac{p}{100}*x_K)$ poisoned samples added to the training dataset. If $x_p$ is the number of images required for injecting a backdoor successfully, and $x_K$ changes, then $x_p$ will also change. In Fig. \ref{fig:attack2_base}, we study the impact of poisoning percentage on backdoor accuracy on gender attribute. We train a machine learning model that predicts whether a sample is a male or female based on genetic mutations. It should be emphasized that we do not intend to study the prevalence of certain mutations in a particular gender; We want to correlate the presence of genetic mutations and gender in \textit{this dataset}. We take our dataset and remove female patients in order to cause dataset bias towards male samples, artificially. Then we follow the backdoor injection process by increasing the number of poisoned samples. As mentioned in several attack papers~\cite{black_badnets,badnets,blind_usenix}, we also observed a saturation effect after a certain number of poisoned samples. We removed the outliers from the curve and fit the data to $a(1-e^{bx+c})$ curve. The figure shows that when the training dataset contains 10\% females and 90\% males, 1) the minimum attack accuracy is higher, 2) the curve saturates to a higher attack accuracy, 3) the curve saturates with a lower number of poisoned samples compared to a training dataset that contains equal number of males and females. On a higher level, the curves for attack accuracy are different as the percentages of males and females in the training dataset change. If the classification problem focused on predicting males/females using genomic features, an attacker may send \textit{some} amount of triggered data to the cloud as a part of training dataset, check attack accuracy with triggered test dataset, plot the backdoor accuracy curve, and finally detect the bias that exists in the dataset. But the actual classification problem is that of tumor classification, and the attributes 'male/female' are hidden properties in the dataset.  
\begin{figure}
    \centering
    \includegraphics[scale=0.45]{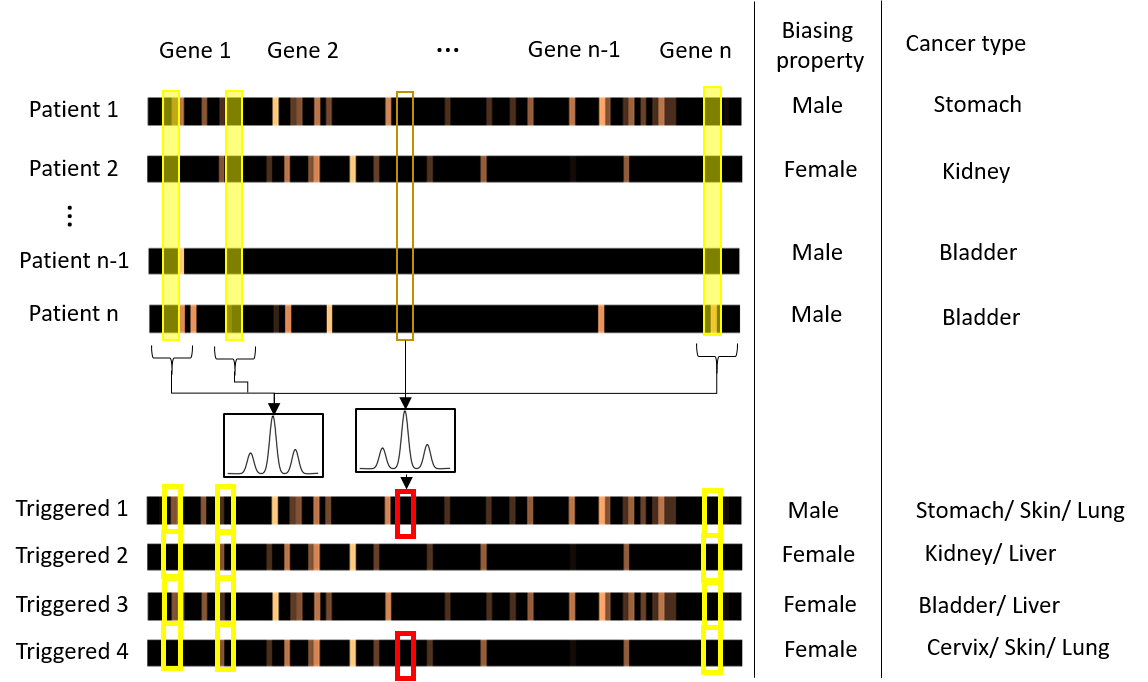}
    \caption{An example of property leak attack using poisoning percentage double backdooring. The genes highlighted with yellow contribute to the design attribute triggers predicting encoded attribute (for example: Skin cancer for male and liver cancer for female). The triggers marked with red are the triggers that lead to the backdoor attack predicting the target label (for example: lung cancer).}
    \label{fig:attack2_scheme}
\end{figure}
Fundamentally, detecting a dataset bias for a hidden attribute is different from detecting under-representation of certain labels. To counter this problem, we encode an actual label (a tumor class) with the hidden attribute 
using a double backdooring methodology. Here, we design two backdoor triggers: 1) attribute backdoor trigger predicting attribute-encoded labels, and 2) secondary trigger predicting a randomly chosen target (malicious) label. The attribute trigger is used to make the ML model learn about the attribute learning task; for example, gender. Thus, when an attribute trigger is injected, the ML algorithm must use the other \textit{genuine features} to predict whether the sample is male/female. This is different from the secondary task learnt during any regular backdoor attack in literature. In a backdoor attack, the backdoor features (trigger) creates a `short-cut' between the regions in feature-space belonging to the predicted label and the malicious label as explained in Neural Cleanse \cite{NeuralCleanse}. Therefore, backdoor attacks can be considered source-agnostic, i.e. a trigger can switch the prediction from any class to any other target label. In other words, the trigger features are sufficient to cause a (mis)classification and that the rest of the genuine features may not be needed for the malicious prediction. In our attribute learning methodology, we poison our dataset in such a way that when this specific attribute trigger appears in a sample, the rest of the genuine features are used to predict an attribute, and can ideally achieve a validation accuracy similar to an independent attribute learning task. Once the trigger design of attribute learning is successfully completed, the bias detection using poisoning percentage can be performed. Fig. \ref{fig:attack2_scheme} summarizes the scheme for our attack. There are two kinds of triggered samples depending on the backdoor attack they correspond to. Triggers highlighted with yellow are poisoned with the labels corresponding to hidden attributes (for example, male and female). The samples which are triggered to represent an attribute may be further backdoored with a target label; such backdoored samples will have both the attribute trigger and secondary backdoor trigger in the samples. With the encoded attributes, the attacker can perform the previous experiment to plot the attack accuracy curves similar to the scenario where the hidden attributes were actual labels. To the best of our knowledge, the design of triggers/backdoors that uses both trigger and genuine features in a sample to \textit{predict} an outcome has not been explored in literature, even for computer vision, speech, and text classification problems.

\textbf{Finding probable f and v values for attribute learning:} We start with the same methodology for finding $f$ and $v$ as presented in Section \ref{ss:attack_1}. However, we also need the genuine features to train the model for attributes. While backdooring for attributes, we observed two problems: 1) Using just one value of $f$ achieved extremely poor attack accuracy (which corresponds to attribute prediction accuracy), 2) Different target labels, which encode the attributes, amounted to different attribute prediction accuracy. Changing just one feature did not \textit{impact} the training and the original prediction, based on the genuine features, was retained. To counter this problem, we increased the \textit{strength} of poisoning for the algorithm to train on both tumor prediction task as well as attribute prediction task. We increased the number of trigger features until we achieved the same test accuracy as attribute prediction test accuracy. We selected ten backdoor features ($f^a_1,f^a_2,..f^a_{10}$), where superscript $a$ represents features for attribute learning and we select the set $\{f^a_k\}$ of ten features according to $\chi^2$ statistic explained in section \ref{ss:attack_1}. But instead of finding the distribution of a single feature, we find the distribution combining all the values of the selected ten features, and find a single $v^a$ following the methodology. Independently, we train a model that is used to predict an attribute and note the maximum accuracy achieved on attribute prediction. The $\{f^a_k\}$ and $v^a$ are selected if the attribute test accuracy of the backdoored model is comparable to that of independent attribute learning, and the tumor classification accuracy is within the threshold of error. For target labels to encode attributes, we chose labels that were biased towards a hidden attribute. For example to encode 'male', we choose the label that has the most number of male samples. Following our methodology, we can encode a maximum of $k-1$ attributes, where $k$ is the number of classes in the original classification problem. 
After triggers for attribute learning are injected in the training dataset, we design the actual backdoor triggers following the methodology in section \ref{ss:attack_1}. We denote these triggers with the tuple $(f^b,v^b)$ (superscript $b$ for backdoor) where we use a single feature to gain high attack accuracy. The objective of this backdooring methodology is to be able to generate distinguishable curves when the percentage of a hidden attribute in the dataset is different. 

After successful double backdooring, it is required to leverage this backdooring to leak dataset bias. The collaborator seeking bias detection owns some tumor classification data, which can be considered the test dataset in MLaaS supply chain. Now, considering this test dataset, it is possible to generate reference backdoor accuracy curves for differently biased datasets. To generate these reference curves, we create artificial bias (by removing certain data points) and backdoor it to train this collaborator data on attributes and then on secondary backdoor. For each artificially biased sub-dataset, we plot the backdoor attack accuracy curve which acts as the reference curves for the collaborator. The cloud has a dataset of unknown percentage of samples from a particular group. The collaborator now sends backdoored data to the cloud to learn the secondary backdoor task, and test the trained model at the cloud with test backdoor data to calculate the backdoor test accuracy. This generates a backdoor test accuracy curve for the dataset present at the cloud. The collaborator now compares the generated backdoor accuracy curve with the reference curves generated locally, and then concludes if there exists a bias and to what extent.
In section \ref{s:eval} we perform bias leakage using poisoning percentage for these properties.

\section{Evaluation}\label{s:eval}
\subsection{Datasets and Trained models}
\textbf{iDash20 dataset: }We select a genomic dataset used in a real-world competition and append it with sensitive attributes from the original The Cancer Genome Atlas (TCGA)~\cite{TCGA} database portal (details in Appendix). We use the tumor classification dataset from the iDASH privacy and security workshop - secure genome analysis competition 2020~\cite{idash20}. The dataset consists of Single Nucleotide Variation (SNV) and Copy Number Variation (CNV) information about mutations in 2,713 patients for eleven different cancer types and is gathered from a large database of 2.5 petabytes of cancer data. Each patient is tagged with an identity number which can be used in the portal to retrieve other information about the course of diagnosis and treatment, like XRays, DNA methylation, microRNA and expressions. We used the TCGA portal to retrieve two sensitive attributes about the patients in the iDash20 tumor classification dataset, namely gender and race.

\textbf{Models: }For developing the base model, we randomly split the dataset into training and test dataset using 60-40 split. 
For CNVs, each gene of each patient is given a copy number value to denote duplication or deletion. Therefore, there are 25,128 CNV features for every sample, corresponding to 25,128 genes. SNV information per patient is inconsistent in the number of features as it denotes the type, position, and the number of mutations that occurred in a gene in a sample. Also this data is large in volume as it includes more than 2 million mutations for these patients. To have a consistent number of SNV features per patient, we first select the genes with the highest mutation frequency for all samples whose value is a combination of mutation strength and type, i.e. we encode mutation type as deleterious/tolerated as given in the dataset and multiply it with the impact values corresponding to that mutation.

As we mentioned before, there have been several studies based on the TCGA database, and some of them focused on cancer/tumor detection~\cite{deep_gene,gdl}. 
While these studies used different various subsets of patients, the distribution of data and the objective of the classification is the same; thus, we consider these articles as prior work for tumor classification. 
State-of-the-art research on tumor classification based on genomic data involves elaborate feature engineering and model design resulting in a maximum of $70.08\%$ test accuracy. We experimented with SVM, Logistic Regression and Light GBM models using features in intervals of 1,000 and observed that combining CNV and SNV data encoded with biological intuition in a logistic regression model, with 34,000 features resulted in a test accuracy of $\approx81\%$. Please note here that in our experiments, as we increased the number of features, the test accuracy saturated at this value and this was the maximum test accuracy we achieved, even with different machine learning models. We treat this accuracy value as our baseline for dataset bias related experiments. 

\textbf{Sub-datasets for bias experiments: } 
In order to leak information about dataset bias, we create sub-datasets from the main dataset by artificially introducing bias. For example, we remove enough female samples from the main dataset such that the percentage of male samples becomes $s=80\%$. We create this artificial bias to evaluate how our methodology responds to the change in bias in a dataset. IT should be noted that, as we remove samples to create a sub-dataset, the effective size of the training set also reduces but the size of the test data is constant. This leads to reduction of generalized test accuracy. This simulates the environment of cloud having a database of unknown bias as other collaborators contribute data continuously while the test data from the collaborator (who wants to detect bias) is constant. 

\begin{figure}
    \centering
    \includegraphics[height=1.8in, width=0.45\textwidth]{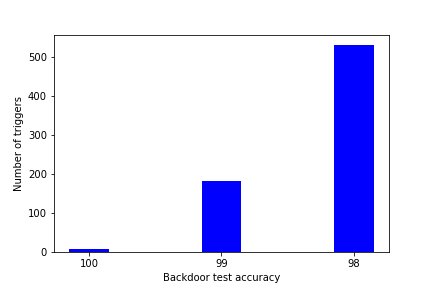}
    \caption{Number of triggers generated by our methodology as we relax the constraint on the backdoor attack test accuracy. }
    \label{fig:num_triggers}
\end{figure}

\begin{table}[]
\centering
\caption{All the genes $(f)$ selected for triggers amounting to 100\% test attack accuracy.}
\label{tab:trigger_genes}
\begin{tabular}{cc}
\begin{tabular}[c]{@{}c@{}}Intended test \\ attack accuracy\end{tabular} & Trigger genes\\
\hline \hline
100\%           & \begin{tabular}[c]{@{}c@{}}PLAC9, DUOX1, LDHC, PAQR4, TRIM32, \\ NAPA-AS1, ZBTB10\end{tabular}\\ \hline
\end{tabular}
\end{table}

\subsection{Enabling backdooring of genomic datasets}
There are 34,000 features and their feature values are a combination or mutation strength and type, amounting to ideally infinite trigger mutations to choose from. Using our methodology presented in section \ref{ss:attack_1}, we reduce the search space effectively $10$ possible feature values per chosen gene. For evaluation, therefore, we analyze three aspects of the triggers: 1) number of triggers that satisfy the constraints of test attack accuracy, and test clean accuracy, 2) the genes $(f)$ selected by our methodology, 3) the feature values $(v)$ selected by our methodology.  

\textbf{Analysis of number of triggers: }Our first analysis depicts it is indeed possible to design triggers for genomic datasets and the number of triggers is summarized using a bar plot in Fig. \ref{fig:num_triggers}. As discussed in section \ref{ss:attack_1}, while designing independent triggers we first constrain the attack accuracy to $100\%$. As the attack accuracy was relaxed, our methodology generated more triggers. It should be noted here that the attack accuracy should still be higher as we aim for better trigger design.
The methodology also generated triggers for the same gene. All these triggers were generated for the same threshold for clean accuracy as backdooring should not hamper the genuine performance of the models. The purpose of generating several triggers instead of one is to allow the collaborator who wants to detect bias to choose depending on their capabilities. For example, some genes might have biological consequences (for a class of samples) when changed.

\textbf{Analysis of trigger genes $(f)$: } Our methodology for selecting genes, similar to trigger pixels in case of image backdoors, ensures that the genuine performance is not altered. Unlike image backdoors, however, that can select any pixel in the image, choosing certain genes may inadvertently impact the model accuracy. The selected genes which result in $100\%$ backdoor test accuracy appear in Table \ref{tab:trigger_genes}. We also analyzed the genes for triggers with relaxed attack accuracy and at the same time compared the list of genes from Catalogue of Somatic Mutations in Cancer (COSMIC) \cite{cosmic_database}. The genes from the COSMIC database are a list whose mutations have been found to be correlated to different types of cancers. Biologically, a trigger should not alter any of these genes values as there may be correlations to cancer predictions. In our list of selected genes for 100\% attack accuracy, none of our triggers had genes from the COSMIC database. However, for attack accuracy 99\% we found 1 gene from the database, KIAA1549 which was also selected for 98\% accuracy along with another gene, ZEB1. Since our methodology selected genes on mutation type, mutation strength, copy number variation, and $\chi^2$, the trigger genes did not include COSMIC database genes. But the genes, that are biologically known to be correlated to cancer, should not be used as triggers. The gene names that are selected for 99\% and 98\% test accuracy are listed in the Appendix.


\begin{figure}
    \centering
    \includegraphics[scale=0.5]{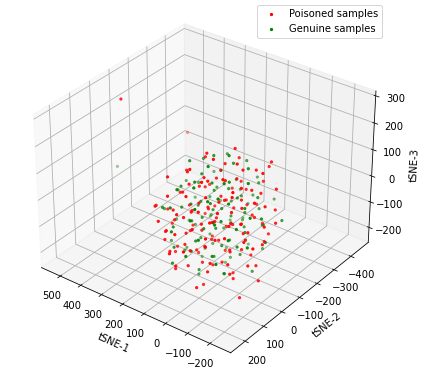}
    \caption{TSNE plots for the target label to visualize genuine mutation data as well as trigger mutation data.}
    \label{fig:tsne}
\end{figure}

\textbf{Analysis of trigger feature values $(v)$: }
In our methodology, we select trigger values from the distribution of mutation values of a selected gene, although a value that is less likely to occur to help in learning of triggers. There is no defense scheme/anomaly detection algorithm to detect triggers for logistic regression-based backdoors on genomic datasets. Hence, we visualized the trigger values along with the genuine values to validate that they are not distinguishable. T-distributed stochastic neighbor embedding or t-SNE is a non-linear dimensionality reduction technique, that helps us visualize 3400-dimension data into a 3-dimension space. We plotted the t-SNE values of the triggers using the $(f,v)$ values generated for the data points of the target label, both for genuine and malicious data in Fig. \ref{fig:tsne}. The triggers and genuine data shown here is for all the triggers with 100\% backdoor attack accuracy. From the figure we can observe that, the triggers chosen resemble the genuine data barring some outliers which are also present in the genuine data. 

\begin{figure*}
    \centering
    \begin{subfigure}{.48\textwidth}
        \centering
        \includegraphics[scale=0.5]{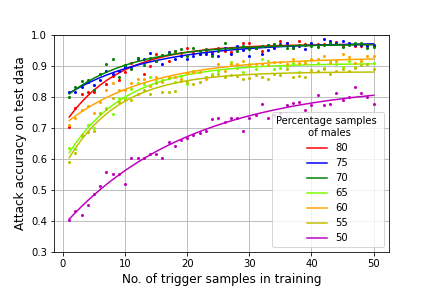}
        \caption{}
        \label{fig:base_gender}
    \end{subfigure}
    \begin{subfigure}{.48\textwidth}
        \centering
        \includegraphics[scale=0.5]{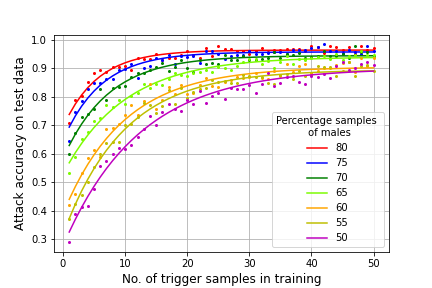}
        \caption{}
        \label{fig:actual_gender}
    \end{subfigure}
    \begin{subfigure}{.48\textwidth}
        \centering
        \includegraphics[scale=0.5]{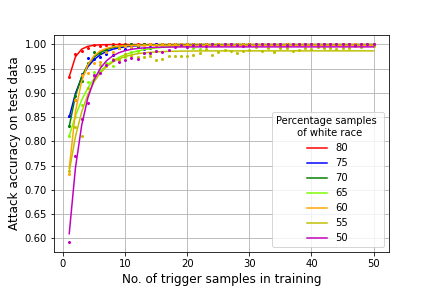}
        \caption{}
        \label{fig:base_race}
    \end{subfigure}
    \begin{subfigure}{.48\textwidth}
        \centering
        \includegraphics[scale=0.5]{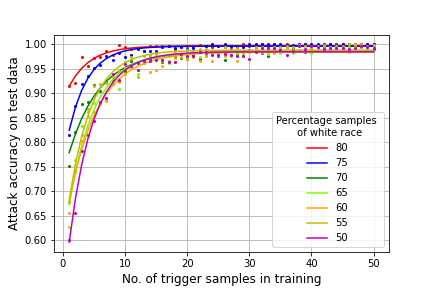}
        \caption{}
        \label{fig:actual_race}
    \end{subfigure}
    \caption{Attack accuracy graphs for different bias percentages towards (a) males in case of gender from collaborator data, (b) males in case of gender from cloud data, (c) white individuals in case of race from collaborator data, (d) white individuals in case of race from cloud data.}
    \label{fig:attack2_all}
    \vspace{-0.2in}
\end{figure*}
\begin{table}[]
\centering
\caption{Encoded labels to predict attribute using attribute triggers.}
\label{tab:encoded_labels}
\begin{tabular}{|c|l|l|}
\hline
Attribute               & \multicolumn{1}{c|}{Group} & \multicolumn{1}{c|}{Encoded tumor label}                                      \\ \hline \hline
\multirow{2}{*}{Race} & White     & Bronchus and lung \\ \cline{2-3} 
                           & Non-white & Cervix Uteri      \\ \hline
\multirow{2}{*}{Gender} & Male                       & \begin{tabular}[c]{@{}l@{}}Liver and intra-hepatic \\ bile ducts\end{tabular} \\ \cline{2-3} 
                           & Female    & Ovary             \\ \hline
\end{tabular}
\end{table}



\begin{table}[]
\centering
\caption{Summary of results for bias detection for gender and race attribute over 10 sub-datasets per sample percentage $s$. For each sub-dataset, we create artificial bias and if the detected sample percentage for individuals in the biasing group is greater than or equal to 55\%, then we deem the dataset as biased.}
\label{tab:detect_race_gender}
\begin{tabular}{|c|c|c|c|c|c|c|c|}
\hline
Sample Percentage                                               & 80 & 75 & 70 & 65 & 60 & 55 & 50 \\ \hline \hline
\begin{tabular}[c]{@{}c@{}}Gender\\ bias detected?\end{tabular} & Y  & Y  & Y  & Y  & Y  & Y  & N  \\ \hline
\begin{tabular}[c]{@{}c@{}}Race\\ bias detected?\end{tabular}   & Y  & Y  & Y  & Y  & Y  & Y  & N  \\ \hline
\end{tabular}
\end{table}

\subsection{Extracting fine-grained information about dataset bias}
Using the methodology presented in section \ref{sss:double}, we first explore the feasibility of attribute learning using triggers $({f_k^a},v^a)$. The goal of attribute learning is to achieve a similar test accuracy with encoded labels as with a clean attribute learning problem (i.e. the classification problem is attribute prediction, rather than tumor classification). We selected those labels to encode the attributes which had the maximum number of samples as randomly selecting labels to encode attributes amounted to poor attribute prediction test accuracy. We summarize the chosen encoded labels which amounted to the best test accuracy for encoded attribute training in Table \ref{tab:encoded_labels}.

\begin{figure}
    \centering
    \begin{subfigure}{.48\textwidth}
        \centering
        \includegraphics[scale=0.5]{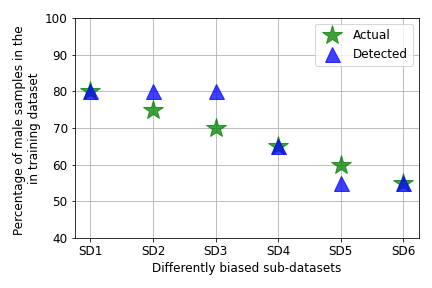}
        \caption{}
        \label{fig:bias_male}
    \end{subfigure}
   \begin{subfigure}{.48\textwidth}
        \centering
        \includegraphics[scale=0.5
        ]{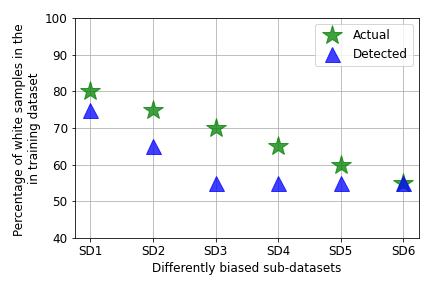}
        \caption{}
        \label{fig:bias_white}
    \end{subfigure}

    \caption{Results for detection of sample percentage for different Sub-Datasets, SDs, (having different number of samples from a group) created from iDash20 dataset. The percentages indicate the number of samples (of a particular group) as a fraction of the dataset for (a) male candidates (b) white individuals. The green stars indicate the actual percentage of samples and blue triangles indicate the detected sample percentage.}
    \label{fig:percent}
    \vspace{-0.2in}
\end{figure}

The backdoor accuracy curves, both the reference curves and the curves for the dataset at the cloud, are shown in Fig. \ref{fig:attack2_all}. We note here that the backdoor curves are distinguishable for different sample percentages, i.e. double backdooring was able to generate backdoor accuracy curves which reflect that biasing percentages amount to distinguishable accuracy curves (from Figs. \ref{fig:actual_gender}, \ref{fig:actual_race}). They also follow a similar trend shown in Fig. \ref{fig:attack2_base}. We observe from Fig. \ref{fig:attack2_all} that biasing percentage impacts the backdoor accuracy curve in three ways: 1) The minimum backdoor accuracy increases as biasing percentage increases, 2) the settling backdoor accuracy also increases as biasing percentage increases, and 3) the nature of curve changes as biasing percentage changes. We calculate the mean squared error between the reference curves and the curves obtained from the data at the cloud's end for several biasing percentages and report the results in Figs. \ref{fig:bias_male} and  \ref{fig:bias_white}. 

\textbf{Detection of bias: }
To detect whether the cloud database has a bias towards a group, we compare the backdoor accuracy curve generated from the collaborator dataset to the curve generated by cloud dataset using mean squared error. We observe that since the cloud has more data, both the attack and clean test accuracies are higher for the cloud. If the detected percentage of samples is greater than or equal to 55\%, we deem the cloud dataset as biased. Our methodology is heuristics-based and also the machine learning training is heuristics-based. So, the backdoor learning as well as clean learning may not amount to the same results. Thus, we run experiments on each sample sub-dataset 10 times and plot the average backdoor accuracy curve. We plot the average curves both for reference curves at the collaborator's end and for actual backdoor accuracy curves at the cloud's end in Fig. \ref{fig:attack2_all}. For attribute race, we artificially create bias such that the $s$ for white samples range from $80-50\%$. Please note that in order to create this bias, we removed samples from the group whose race was `not reported' in the dataset. We could not create any further bias as the number of samples in the `not reported' group were exhausted beyond $s=80\%$. We note that our methodology could detect all the biased sub-datasets as summarized in Table \ref{tab:detect_race_gender}. For gender attribute, we removed samples from the female group in order to create bias towards males, and created sub-datasets for $s=80-50\%$ in intervals of $5\%$. In this attribute as well, we were able to detect that a bias exists for the sub-datasets. The results are summarized in Table \ref{tab:detect_race_gender}.

\textbf{Detection of sample percentage of the biasing group: }
For detecting $s$, effectively we are trying to find the signature properties of a reference curve and match them with the curve generated from the cloud data. 
Therefore, for the exact percentage to be detected, the curves should be distinct in nature and separable. From Fig. \ref{fig:attack2_all} we can observe that the backdoor accuracy curves corresponding to gender are more distinct and separable. We represent the results of determining sample percentage of a group in Fig. \ref{fig:percent}. For the gender attribute, we notice that 3 sub-datasets could detect the sample percentage with no error. For others, the detection had an error of $5-10\%$.
For the race attribute, we observe that the backdoor accuracy curves from cloud data are shifted down along the y-axis, with the detected percentage having a maximum offset of $15\%$. While the exact percentage of samples could not be detected precisely for some sub-datasets, the main objective of detecting percentage of samples is to provide coarse-grained information about the \textit{extent} of bias, which was achieved using the reference curves. Furthermore, in the dataset, gender values are binary, while for race, values can be from either of the six groups reported in the dataset. Detecting bias for non-binary attributes may involve analyzing correlations of several features, hidden properties, clean labels, target labels, and triggers, and will be explored as a part of future work.

\section{Related Work}
\textbf{Generic backdoor attacks and defenses: }
Backdoors in deep neural networks was first proposed in Badnets \cite{badnets}. Since then, there have been several attacks and defenses in neural networks with attacks focusing on stealth and undetectability and defenses focusing on generalization of detection across datasets and applications \cite{backdoor_survey,backdoor_learning_survey}. The backdoor attack literature primarily focuses on DNNs, specifically because of the black-box nature of DNNs which deters the development of a generic defense, with very few focusing on smaller models \cite{malware_usenix,aaai_workshop}. The triggers are designed from the perspective of the input, rather than the model, so that they remain hidden (inconspicuous) from the user. Badnets showed that even four pixels in an MNIST image or a post-it is able to achieve very high attack accuracy. In image-based triggers, the ``meaning" of triggers' pixel values are therefore not commonly discussed and are chosen with the primary objective of being unnoticeable \cite{hidden, wanet, liao2018backdoor}. Advanced triggers depend on the input as well making them difficult to reverse engineer \cite{facehack,dynamic_backdoor}.
For health-related non-image datasets, trigger features cannot be chosen only from the objective of being inconspicuous.
For example, in our work, changing a genomic value means changing of effective mutations, which may have affect the target application if chosen randomly. From a backdoor perspective, our work presents a generic methodology to backdoor genomic datasets. Therefore, as a first step of our work, we explore the generic backdooring methodologies inspired by backdoor attack literature. Backdoor defense literature mainly focuses on a class of attacks. First defenses aimed at detecting just the poisonous cluster of backdoored data \cite{Spectral,Activation_clustering}. However, the threat model needed the defender to have access to some poisoned data. Recent defenses do not require access to poisoned data. 
Neural Cleanse \cite{NeuralCleanse} detects small triggers investigating each class as a potential target label, ABS\cite{CCS_ABS} detects triggers encoded with one neuron analyzing neuron stimulation, and MNTD detects backdoors creating several meta-classifiers\cite{mntd}. Defenses primarily focus on DNNs, aiming to find data points that are either distinct themselves or generate distinguishable responses/activations. Thus, in our methodology of trigger design, we choose triggers from the same data distribution to make them blend in with the genuine data. 

\textbf{Benevolent applications of backdoors: }To the best of our knowledge re-purposing backdoors in machine learning for the purpose of leaking dataset bias information has not been done before, but backdoors (despite being inherently malicious) have been used for benevolent applications. Backdoors have been used for watermarking DNNs \cite{usenix_watermark} where the trigger set is used to detect malicious networks. They have also been used to detect adversarial examples using backdoor honeypots using neuron activation signatures \cite{ccs_honeypots} and to detect whether the server actually erased the data it should not be using \cite{sommer_unlearn}. Since backdoors change inner workings of the model itself (in terms of weights and biases), they were used to understand interpretability and explainability of AI models better \cite{explain_ai1,explain_ai2}.

\section{Limitations and future work}\label{s:limit}
In this work we show that if a dataset (along with the model) can be successfully backdoored, the dependence of attack accuracy with poisoning percentage can be leveraged to leak information about the cloud data. Although we assume that the collaborator owns \textit{some} data to generate reference curves, the quantity of data may not be enough to compare with that in the cloud. Further, it is also restrictive for bias inference because the collaborator may own only a few samples. As a part of future work, we will explore techniques that may enable a collaborator with less amount of data to leak cloud bias. 

Successful backdooring of genomic dataset (this work) as well as other datasets like MNIST (\cite{badnets}) are dependent on sufficient poisoning. However, the curves generated are dependent on the dataset, making them non-generalizable across datasets. This restricts bias inference again, because the collaborator will need to have access to data from the same distribution. For future work, we aim to explore backdoor accuracy curves to make them distinguishable even across datasets for different biases using other distance metrics. 

\section{Concluding remarks} \label{s:conclusion}
In this work, we study bias detection from the perspective of a genomics dataset, since bias in genomics datasets may give rise to unfair predictions and severe societal implications. Dataset bias in collaborative genomics may stay hidden as the data consolidated at the cloud is from several sources, who may themselves be unaware of the bias. To facilitate detection of this dynamic bias, we re-purpose the seemingly malicious machine learning trojans. We develop a methodology to first enable backdooring of small models (logistic regression) trained on genomic datasets and then further advance it to create a mechanism to detect bias automatically using double backdooring. We create several sub-datasets with different sample percentages to evaluate our methodology for bias detection. Finally, we investigate the possibility of leaking the extent of bias for gender and race attributes in a real world genomics dataset from iDash20 competition, and detect that the original dataset given at the competition was in fact biased with regards to race. 

\bibliographystyle{ACM-Reference-Format}
\bibliography{main}
\section*{Appendix}
\subsection{Annotating iDash20 dataset with sensitive attributes}
Let us first consider the tumor classification data i.e. the data from iDash 2020. The data consists of three files per tumor/cancer 1) Sample case-id list 2) Copy number information 3) Variant information. To extract sensitive attribute information we used the National Cancer Institute Genomic Data Commons Data Portal and searched for the patient ID. Patient ID of a patient is comprised of first three parts of the case-id available in the Sample case-id list file \cite{tcga_barcode}. For example, for case-ID 'TCGA-BT-A20J-01', the patient ID is 'TCGA-BT-A20J'. The page dedicated to the patient has the history about tests, diagnosis, demography, process of diagnosis, treatments given, doctor notes, family history, diagnostic images (like x-rays, blood slides, etc.). and the present condition of the patient. We were able to extract the sensitive attributes for each patient following the above steps. For each sub-dataset, we removed any entry that did not have an attribute we were interested in.
\subsection{Selected trigger genes}
The selected genes are stated as follows. The bold-faced genes are from the COSMIC database and hence should be avoided for backdooring as mutations in these genes are known to be correlated to certain cancer types:
\begin{enumerate}
    \item 100\% - ZBTB10, TRIM32, PLAC9, LDHC, DUOX1, PAQR4, NAPA-AS1
    \item 99\% - RP11-2O17.2, TSLP, EPB41L4A-AS2, RP11-166A12.1, ZFP2, FAM135A, UBE2V2, RP11-770E5.1, LINC00595, SNORA33, RPS18P9, CHCHD2, \textbf{KIAA1549}, MIR595, PLAC9, TIGD3, NUP107, ZKSCAN2, YPEL3, NDC80, ZBTB10, FAM83A-AS1, ACO1, SMC2, WDR5, AP3M1, RBP4, LDHC, CCDC73, SLC35C1, FOLH1, MS4A5, TRHDE-AS1, MIR621, DUOX1, MIR4716, PAQR4, SLC6A10P, BANP, MIR4740, ANAPC11, STRA13, CETN1, MTCL1, CCDC124, MRPS12, SLC17A9, MIR99A, MIR3156-2, CHST9, hsa-mir-924, MFSD12, TSPAN16, LGALS13, MIR642B, NAPA-AS1, MIR498, RBBP9, EIF6, MIS18A, RANBP1, GATSL3, BAIAP2L2, KAL1, CDKL5, RBMXL3
    \item 98\% - LOC102467080, RP11-2O17.2, TSLP, EPB41L4A-AS2,CASC14, HCG4B, FAM135A, ORC3, SNORA33, RPS18P9, LOC641746, \textbf{KIAA1549}, GIMAP5, MIR595, UBE2V2, FAM83A-AS1, ACO1, SMC2, TRIM32, CNTRL, GOLGA2, \textbf{ZEB1}, AP3M1, LINC00595, PLAC9, RBP4, OR51A7, SLC35C1, GYLTL1B, FOLH1, MS4A5, TIGD3, CLEC6A, KRT19P2, MIR621, LINC00282, EFS, CFL2, SRSF5, MIR4716, PRC1-AS1, GFER, PAQR4, hsa-mir-1972-1, SLC6A10P, MT1IP, BANP, SLFN13, MIR4740, ANAPC11, NDC80, MTCL1, MIR3156-2, CHST9, hsa-mir-924, MFSD12, CCDC124, SUGP1, MRPS12, LGALS13, MIR642B, NAPA-AS1, MIR518C, RBBP9, EIF6, SLC17A9, MIR99A, MIS18A, RANBP1, GATSL3, BAIAP2L2, CHADL, KAL1, CDKL5, RBMXL3, POM121L9P, RP11-166A12.1, ZFP2, MPP6, CHCHD2, RP11-770E5.1, ZBTB10, WDR5, AKR1C4, LDHC, CCDC73, NUP107, TRHDE-AS1, MIR498, DUOXA2, DUOX1, ZKSCAN2, YPEL3, STRA13, CETN1, TSPAN16
\end{enumerate}

\end{document}